\documentclass[11pt]{article}

\usepackage{acl}

\usepackage{times}
\usepackage{latexsym}

\usepackage[T1]{fontenc}

\usepackage[utf8]{inputenc}

\usepackage{microtype}

\usepackage{hyperref}
\usepackage{amsmath, amsfonts, amsthm}
\usepackage{mathtools}
\usepackage{autobreak}
\usepackage{tabularx}
\usepackage{arydshln}
\usepackage{tikz}

\usepackage{algpseudocode}
\usepackage{algorithm}

\usepackage{comment}

\theoremstyle{definition}
\newtheorem{definition}{Definition}

\newcommand{\argmax}{\mathop{\rm arg~max}\limits}

\title{Why Guided Dialog Policy Learning performs well?\\
Understanding the role of adversarial learning and its alternative}

\author{
    Sho Shimoyama${}^{1}$, Tetsuro Morimura${}^{2}$, Kenshi Abe${}^{2}$, Toda Takamichi${}^{1}$, Yuta Tomomatsu${}^{1}$, \\
    {\bf Masakazu Sugiyama${}^{1}$, Asahi Hentona${}^{2}$, Yuuki Azuma${}^{1}$, Hirotaka Ninomiya${}^{1}$}\\
    ${}^{1}$AI Shift Inc.,
    ${}^{2}$CyberAgent Inc. \\
  \texttt{shimoyama\_sho@ai-shift.jp},
  \texttt{\{morimura\_tetsuro, abe\_kenshi, }\\
  \texttt{toda\_takamichi, tomomatsu\_yuta, sugiyama\_masakazu, hentona\_asahi,}\\
  \texttt{azuma\_yuki, ninomiya\_hirotaka\}@cyberagent.co.jp}
}

\begin{document}
\maketitle

\begin{abstract}
Dialog policies, which determine a system's action based on the current state at each dialog turn, are crucial to the success of the dialog.
In recent years, reinforcement learning (RL) has emerged as a promising option for dialog policy learning (DPL).
In RL-based DPL, dialog policies are updated according to rewards.
The manual construction of fine-grained rewards, such as state-action-based ones, to effectively guide the dialog policy is challenging in multi-domain task-oriented dialog scenarios with numerous state-action pair combinations.
One way to estimate rewards from collected data is to train the reward estimator and dialog policy simultaneously using adversarial learning (AL).
Although this method has demonstrated superior performance experimentally, it is fraught with the inherent problems of AL, such as mode collapse.
This paper first identifies the role of AL in DPL through detailed analyses of the objective functions of dialog policy and reward estimator.
Next, based on these analyses, we propose a method that eliminates AL from reward estimation and DPL while retaining its advantages.
We evaluate our method using MultiWOZ, a multi-domain task-oriented dialog corpus.
\end{abstract}

\section{Introduction}
Dialog policy, which chooses a system action for a user action based on the current state, is an essential component of pipeline-based multi-domain task-oriented dialog systems~\citep{convlab2, survey_recent_rl_dialog_system}.
In recent years, reinforcement learning (RL) methods have been widely used for dialog policy learning (DPL)~\citep{peng-etal-2017-composite, Peng2018DeepDynaQ,cao-etal-2020-adaptive, takanobu-etal-2020-multi, Zhang_Zeng_Lu_Wu_Zhang_2022-efficient-dialog}.
In RL-based DPL, the dialog policy is learned through user interactions and assessments.
As RL requires numerous interactions, user simulators mimicking real user behavior are often used in place of real users~\citep{wang-etal-2020-learning-efficient,wang-wong-2021-collaborative,lin-etal-2022-gentus}.
In such cases, the reward function evaluates the quality of the dialog.

Although the reward function is essential for RL-based DPL, its manual construction is difficult.
The typical reward function, which gives a small negative reward at each turn and a large positive reward at the end of the dialog on its successful completion, is unsuitable for DPL in task-oriented dialogs.
In task-oriented dialogs, short-turn dialogs resulting from negative rewards at each turn are not always appropriate.
When no entity satisfies the user's constraints, offering alternatives rather than terminating the dialog may be preferable. 
In addition, the above reward function slows down the learning of the dialog policy due to the absence of an evaluation of good or bad actions at each turn.
This is known as the sparse reward problem in RL \citep{pmlr-v80-riedmiller18a,  rengarajan2022reinforcement}.
One way to overcome this problem is to make state-action-based rewards rather than session-based ones.
However, manually designing such rewards is challenging in multi-domain task-oriented dialog scenarios, where there are numerous state-action pair combinations.

To address the above issues, Guided Dialog Policy Learning (GDPL)~\citep{takanobu-etal-2019-guided} combines adversarial learning (AL)~\cite{goodfellow-adversarial} and inverse reinforcement learning (IRL)~\cite{ziebart-max-ent} to estimate rewards from the collected data.
In GDPL, the reward estimator is a state-action-based function.
Following the concept of AL, GDPL alternately trains the dialog policy and the reward estimator, where the dialog policy is updated to maximize the expected cumulative rewards.
The reward estimator is updated to give high values for state-action pairs in expert dialog sessions and low values for ones in policy-generated sessions.
GDPL has been shown to outperform a method using AL to train a session-based reward estimator in \citet{takanobu-etal-2019-guided}.
This indicates that state-action-based rewards are more efficient than session-based rewards for DPL.
However, it remains unclear how AL affects the training of dialog policy in GDPL. 
Moreover, AL causes specific problems, such as mode collapse~\citep{veegan}, which degenerate the performance of the dialog policy.

In this study, we analyze the effects of AL on the learning of dialog policy in GDPL.
We found that one of the effects of AL is to prevent the same state action from being repeated in one dialog session through analyses for the objective functions of the reward estimator and the dialog policy.
Moreover, to address AL-specific problems, we propose a reward estimation method that eliminates AL while retaining its advantages.
We maintain the above effect by assigning a large negative reward to repeated state-action pairs, thereby constructing a reward function without using AL.
The usefulness of our method is verified through experiments with MultiWOZ~\cite{budzianowski-etal-2018-multiwoz}, a multi-domain task-oriented dialog corpus, and an agenda-based user simulator~\cite{schatzmann-etal-2007-agenda}.
Our contributions are as follows:
\begin{itemize}
    \item We identify one of the roles of AL in GDPL based on the analyses of the objective functions.
    \item We propose a reward estimation method without using AL. This method is as effective as those that use AL and avoids an AL-specific problem, known as 'mode collapse'.
\end{itemize}

\section{Related Work}
\subsection{Reward Estimation with Domain Knowledge, Additional Label}

In RL-based DPL, dialog policy is updated based on rewards.
Several reward estimation methods have been proposed to obtain a superior dialog policy, such as hand-crafted rewards, rewards based on domain knowledge, and rewards with additional labels.
The method of giving human feedback for the selected actions at each turn, in addition to a success reward at the end of the session, requires fewer dialog sessions compared to that using only a success reward~\cite{parath-interactive-rl}.
An active learning method with a Gaussian process dialog success predictor~\cite{su-etal-2016-line} decreases the number of user feedbacks required by requesting it only when the dialog sessions' predicted variance is high.
Estimating the weights of multiple handcrafted rewards with multi-objective RL~\cite{ultes-etal-2017-reward}  shows good performance compared to using a single reward.
The method using Interaction Quality, an expert annotation label for dialog quality, as a reward has been studied in \citet{ultes-2019-improving}.
Interaction Quality is a more direct indicator of good or bad dialog than dialog success.
However, these reward estimators require domain knowledge, user feedback, and additional labels, thus making their construction expensive.

\subsection{Reward Estimation with Adversarial Inverse Reinforcement Learning}
IRL estimates a reward function from expert data~\cite{ziebart-max-ent}.
Adversarial IRL, which combines AL with IRL, has demonstrated superior performance in large, high-dimensional settings~\cite{ho-gail,fu2018learning}.
Encouraged by the success of adversarial IRL in other research fields, AL-based reward estimation methods have been proposed for DPL.
In DPL, the discriminator is the reward estimator, and the generator is the dialog policy.
The reward estimator is typically represented by either a binary classifier or a real-valued function.
In \citet{liu-lane-2018-adversarial}, AL is used to alternately train a binary classifier session-based reward estimator and a dialog policy.
The reward estimator is learned to give 0 for the policy-generated sessions and 1 for expert sessions.
In \citet{li-etal-2020-guided},  a binary classifier reward estimator and a dialog policy are trained sequentially rather than alternately.
First, this method trains the state-action-based reward estimator using an auxiliary state-action generator with AL.
Then, the reward estimator is fixed, and only the dialog policy is updated.
By separating the learning of the reward estimator and the dialog policy, this method can mitigate the problem of local optima associated with AL.
Unlike the previous two methods, GDPL~\cite{takanobu-etal-2019-guided} employs a state-action-based reward estimator that generates real-valued outputs.
GDPL learns the reward estimator to give high values for the state-action pairs that appear in expert dialog sessions and low values for ones that appear in policy-generated sessions.
By using state-action-based rewards rather than session-based ones, GDPL deals with the sparse reward problem.
In addition, GDPL adapts to large-space settings such as multi-domain task-oriented dialogs, where state-action pair combinations  explode, using adversarial IRL.

In this work, we focus on the analysis and improvement of the method that uses a real-valued reward estimator (i.e. GDPL).

\section{Preliminary}
In this work, we target a DPL scheme that updates a dialog policy $\pi_\theta$ through interaction with a user simulator $\mu$.
In this scheme, each dialog session $\tau$ with dialog turn $|\tau|$ is a trajectory of state-action pairs $\{(s_0,a_0), \ldots, (s_{|\tau|},a_{|\tau|}) \}$.
At each turn $t$, a dialog state tracker (DST) manages the current state $s_t$ and
the user simulator $\mu(a_t^u|s_t^u)$ selects a user action $a_t^u$ based on the current state $s_t^u$ which is derived from the previous state $s_{t-1}$ and action $a_{t-1}$ using the DST.
Also, the dialog policy $\pi_\theta(a_t|s_t)$ selects a system action $a_t$ based on the current state $s_t$ provided by the DST according to the $s_t^u$ and $a_t^u$.
A reward estimator $f_w$ evaluates the dialog sessions generated by the above interactions and the dialog policy $\pi_\theta$ aims to maximize these evaluation values.

\subsection{Dialog State Tracker, State and Action}
During dialog session, DST~\cite{gao-etal-2019-dialog, lee-etal-2021-dialogue} manages the belief state, containing informable slots which show user constraints and requestable slots which represent requests from users.
The belief state is updated based on the currently selected user and system action. \\
\textbf{Action}: Action $a_t$ is a subset of the dialog act set.
Dialog act~\cite{stolcke-etal-2000-dialogue} is quadruple $\{$domain, intent, slot-type,  slot-value$\}$ (e.g. $\{$ restaurant, inform, price range, moderate $\}$) representing the intent of user or system utterance.\\
\textbf{State}: State $s_t$ is a structured representation of the dialog history up to turn t, consisting of $\{a^u_t, a_{t-1}, b_t, q_t\}$,
where $b_t$ is the belief state from DST at turn $t$ and $q_t$ is the number of query results from an external database.

\subsection{Guided Dialog Policy Learning: GDPL}
GDPL~\cite{takanobu-etal-2019-guided} learns the state-action-based real-valued reward estimator $f_w(s, a)$ and the dialog policy $\pi_\theta$ alternately.
The dialog policy $\pi_\theta$ and the reward estimator $f_w$ are neural networks with parameters $\theta$ and $w$, respectively.
Building on maximum entropy IRL~\cite{ziebart-max-ent}, the dialog policy $\pi_\theta$ maximizes the expected cumulative reward for a generated dialog $\tau$ calculated with the reward estimator $f_w$,
\begin{align}
    \label{eq: obj_policy_gdpl}
    \max_{\theta} \mathbb{E}_{\tau \sim \pi_\theta}\left[\sum_{t=0}^{|\tau|}\gamma^{t} \hat{r}(s_t, a_t)\right], \\
    \hat{r}(s_t,a_t) = f_w(s_t, a_t) - \log \pi_\theta(a_t|s_t). \notag
\end{align}
To distinguish between human-human dialogs and human-system dialogs, the reward estimator $f_w$ aims to assign high values to state-action pairs in expert dialog sessions and low values to those in policy-generated dialog sessions.
This is formulated as the optimization problem:
\begin{align}
    \label{eq: obj_reward_gdpl}
    \max_{w} \mathbb{E}_{(s,a) \sim p_D} \left[f_w(s,a)\right] - \mathbb{E}_{(s,a) \sim p_{\pi_\theta}} \left[ f_w(s,a) \right],
\end{align}
where $p_D(s,a)$ is an empirical distribution constructed from expert data $D$, $p_{\pi_\theta}(s,a)$ is the probability that state-action pair $(s,a)$ occurs in dialog sessions generated by the dialog policy $\pi_\theta$. 
In the above equation, the second term represents the AL term.
GDPL employs a rule-based DST.

\section{Analysis of the Role of Adversarial Learning in GDPL}
In this section, we show that AL prevents the same state-action pairs from repeating unnecessarily in one dialog session.
We refer to such a dialog as a loop dialog and it is defined in Sec.~\ref{sec: def_of_loop_dialog} along with illustrative examples.
In Sec.~\ref{sec: analysis_for_obj_func}, we analyze the role of AL in GDPL.
In particular, we present the following two facts: (1) The dialog policy $\pi_\theta$ frequently generates loop dialogs when using the reward estimator $f_w$ trained without AL. (2) The reward estimator $f_w$ gives large negative values to loop dialogs when using the dialog policy $\pi_\theta$ such as (1).

\subsection{Definition of Loop Dialog}
\label{sec: def_of_loop_dialog}
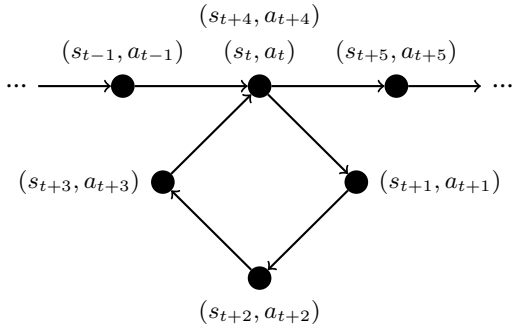
\begin{figure}[htbp]
\centering
\begin{tikzpicture}[node distance={18mm}, thick, main/.style = {draw, circle, fill, inner sep=1mm}]
    \tikzset{
        node/.style = {draw, circle, fill, minimum size=0pt, inner sep=0mm}
        node/.default = 0pt
    }
    \node[main, label=above:{\footnotesize $(s_{t-1}, a_{t-1})$}] (1) { }; 
    \node[main, align=left, label={[align=center]{\footnotesize$(s_{t+4}, a_{t+4})$}\\{\footnotesize$(s_{t}, a_{t})$}}] (2) [right of=1] { }; 
    \node[main, label=above:{\footnotesize $(s_{t+5}, a_{t+5})$}] (3) [right of=2] { };
    \node[main, label=right:{\footnotesize $(s_{t+1}, a_{t+1})$}] (4) [below right of=2] { }; 
    \node[main, label=below:{\footnotesize $(s_{t+2}, a_{t+2})$}] (5) [below left of=4] { }; 
    \node[main, label=left:{\footnotesize $(s_{t+3}, a_{t+3})$}] (6) [below left of=2] { }; 
    \node[node] (7) at (5,0) {...};
    \node[node] (8) at (-1.4,0) {...};
    
    \draw[->] (1) -- (2); 
    \draw[->] (2) -- (3); 
    \draw[->] (2) -- (4); 
    \draw[->] (4) -- (5);
    \draw[->] (5) -- (6);
    \draw[->] (6) -- (2);
    \draw[->] (3) -- (7);
    \draw[->] (8) -- (1);
    
\end{tikzpicture} 
\caption{An illustration of a loop dialog for $n=4$. A loop occurs during the transition of state-action pairs from $(s_t, a_t)$ to $(s_{t+4}, a_{t+4})$. The loop state-action is $(s_{t+4}, a_{t+4})$.}
\label{fig:image_of_loop_dialog}
\end{figure}

Loop dialog includes the same state-action pair multiple times.
Namely, 
\begin{definition}[loop dialog]
We say a dialog $\tau$ is a loop dialog if there is a turn $t$ and step $n > 0$ such that  $(s_t, a_t) = (s_{t+n}, a_{t+n})$ in $\tau$. 
Then we say that $s_{t+n}$ and $a_{t+n}$ are the loop state and action, respectively. 
\end{definition}
Roughly speaking, loop dialogs are the dialogs whose state has not progressed at all for at least $n$ turns.
Figure~\ref{fig:image_of_loop_dialog} is an illustration of a loop dialog for $n=4$ and Table~\ref{tab: loop_dialog} is an example of a loop dialog.
In Table~\ref{tab: loop_dialog}, the same user and system utterances are repeated after turn $t=2$.
In other words, the same user and system actions are continuously chosen.
As a result, the state $s_t$ is un-updated for $t\geq2$ because system action $a_{t-1}$, user action $a_t^u$, the belief state $b_t$, and $q_t$ remain unchanged.

\begin{table}[bth]
    \fontsize{10pt}{12pt}\selectfont
    \centering
    \begin{tabularx}{\linewidth}{c|X}
        \hline
        speaker & utterance \\
        \hline
        usr & Need a restaurant called abc restaurant. \\
        sys & There's a place called abc restaurant. \\
        & I will book it for you and get a reference number?\\
        \hdashline
        usr & I am looking for details on abc restaurant.\\
        sys & Would you like to try abc restaurant?\\
        \hdashline
        usr &  I am looking for details on abc restaurant.\\
        sys & Would you like to try abc restaurant?\\
        \hdashline
        & ... \\
        \hdashline
        usr & I am looking for details on abc restaurant.\\
        sys & Would you like to try abc restaurant?\\
        \hline
    \end{tabularx}
    \caption{An example of a loop dialog within user (usr) and system (sys). After the second turn, the same utterances continue and the dialog state does not progress.}
    \label{tab: loop_dialog}
\end{table}

\subsection{Efficacy of Adversarial Learning for Loop Dialogs}
\label{sec: analysis_for_obj_func}
In this section, we show that AL suppresses loop dialogs by considering the objective function~\eqref{eq: obj_reward_gdpl} for the reward estimator $f_w$ and the one~\eqref{eq: obj_policy_gdpl} for the dialog policy $\pi_\theta$.

First, we show that one of the dialog policies learned with a reward estimator, trained without AL, generates loop dialogs frequently.
We consider the optimization problem in Eq.~\ref{eq: obj_reward_wo_AL}, which is derived by removing the AL term from Eq.~\eqref{eq: obj_reward_gdpl}.
However, the optimization problem is not well defined because there is an undesirable trivial solution that gives $+\infty$ for all state-action pairs $(s, a) \in D$.
\begin{align}
    \label{eq: obj_reward_wo_AL}
    \argmax_{w} \mathbb{E}_{(s,a) \sim p_D} \left[f_w(s,a)\right].
\end{align}
As a result,  a learned reward estimator $f_{\hat{w}}$ based on Eq.~\eqref{eq: obj_reward_wo_AL} probably gives similar high positive values to $(s, a) \in D$.

We suspect that the dialog policy trained with the reward estimator $f_{\hat{w}}$ tends to generate loop dialogs with each state-action pair $(s_t, a_t) \in D$ due to the following reasons:
\begin{enumerate}
    \setlength{\itemsep}{-3mm} 
    \item A learned dialog policy $\pi_{\hat{\theta}}$ based on Eq.~\eqref{eq: obj_policy_gdpl} often produces dialogs with high cumulative rewards, $\sum_{t=0}^{|\tau|}\gamma^{t} \hat{r}(s_t, a_t)$.\\
    \item With the reward estimator $f_{\hat{w}}$, loop dialogs with each state-action pair $(s_t, a_t) \in D$ can achieve high cumulative rewards because:
    \begin{itemize}
        \setlength{\parskip}{1mm} 
        \setlength{\itemsep}{3mm} 
        \item Loop dialog repeats loop state-action pair $(s_{t+n}, a_{t+n})$ (details in~\ref{appendix: loop_dialog_repeat}). 
        This results in a longer dialog turn $|\tau|$.
        In particular, $|\tau| \geq n$.
        \item The value of $f_{\hat{w}}(s_t, a_t)$ is large and positive for each state-action pair $(s_t, a_t) \in D$. 
       As the dialog turn $|\tau|$ is long, the cumulative reward $\sum_{t=0}^{|\tau|}\gamma^{t} \hat{r}(s_t, a_t)$ becomes large for such loop dialogs.
    \end{itemize}
\end{enumerate}
These factors contribute to the frequent generation of the above loop dialogs by the learned dialog policy $\pi_{\hat{\theta}}$.
Next, we show that the reward estimator trained with AL suppresses loop dialogs by assigning large negative values to loop state actions.

The objective function~\eqref{eq: obj_reward_gdpl} of reward estimator $f_w$ is equivalent to the following equation:
\begin{align}
    \max_{w} \sum_{s\in \mathcal{S}} \sum_{a\in\mathcal{A}} f_w(s,a) \left( p_\mathcal{D}(s,a) -  p_{\pi_\theta}(s,a) \right),
\end{align}
where $S$ is the set of all states and $A$ is a set of all actions.
An optimal reward estimator $f_{w^\ast}$ obtained by solving the above equation gives $+\infty$ for the state-action pairs satisfying $p_\mathcal{D}(s,a) -  p_{\pi_\theta}(s,a) > 0$ and $-\infty$ for ones with $p_\mathcal{D}(s,a) -  p_{\pi_\theta}(s,a) < 0$.
$p_\mathcal{D}(s,a) - p_{\pi_\theta}(s,a) < 0$ means that the probability of occurrence of (s,a) in the dialog session generated by $\pi_\theta$ is greater than that in D.
For the dialog policy $\pi_\theta^{\rm loop}$ which frequently generates loop dialogs, 
$p_\mathcal{D}(s^{\rm loop},a^{\rm loop}) - p_{\pi_\theta^{\rm loop}}(s^{\rm loop},a^{\rm loop}) < 0$ probably holds for loop state actions $(s^{\rm loop}, a^{\rm loop})$ because $p_{\pi_\theta^{\rm loop}}(s^{\rm loop},a^{\rm loop})$ is high.
Thus, the learned reward estimator $f_{\hat{w}}$ with $\pi_\theta^{\rm loop}$ assigns large negative values for loop state-actions.
Using the above reward estimator, the updated dialog policy based on Eq.~\eqref{eq: obj_policy_gdpl} reduces the occurrence of loop dialogs because the cumulative rewards for loop dialogs are significantly small.

In this section, we show that AL in GDPL suppresses loop dialogs.
However, AL could cause specific issues, such as mode collapse, which degrade the performance of dialog policy.

\section{Proposed Reward Estimation Method}
Mode collapse, an AL-specific problem, can reduce the performance of dialog policy.
In this phenomenon, the learned model captures only certain trends from the training data, leading to biased output despite the presence of various trends.
Consequently, the dialog policy chooses only a few actions, ignoring other available options, such as reservation actions, even when reservable entities exist.

To avoid AL-specific problems like the above, we propose a simple reward estimation method that removes AL from both the learning of reward estimator and dialog policy.
The main idea is to retain the effectiveness of AL in suppressing loop dialogs without implementing it.

Similar to maximum entropy IRL~\cite{ziebart-max-ent}, we represent the state-action-based reward estimator $p_w(s,a)$ with the Boltzmann distribution.
This is learned by maximizing the log-likelihood of human expert demonstrations:
\begin{align}
    p_w(s,a) &= \frac{\exp(f_w(s,a))}{\sum_{s' \in S, a' \in A} \exp(f_w(s', a'))}, \\
    \label{eq: MLE}
    \max_{w} & \  \mathbb{E}_{(s,a) \sim p_D}[\log p_w(s,a)],
\end{align}
where $f_w$ is a neural networks with parameter $w$ and $p_D(s,a)$ is an empirical distribution constructed from human expert dialog sessions $D$.
Using this reward estimator, the reward function $\hat{r}$ gives a large negative value $-L$ for loop state actions to avoid loop dialogs and positive values for other state-action pairs by subtracting $f_{\text{min}} = \min_{(s,a)\in S \times A} \log p_w(s,a)$ to avoid excessive dialog turn shortening,
\begin{equation}
    \hat{r}(s,a) = \begin{cases}
        -L \\
        \log p_w(s,a) - f_{\text{min}} - \log \pi_\theta(a|s)\\
    \end{cases}.
\end{equation}

In maximization problem \eqref{eq: MLE}, calculating the partition function $\sum_{s' \in S, a' \in A} \exp(f_w(s', a'))$ becomes challenging due to the explosion of state-action combination.
In our implementation setup, the number of combinations is approximately $2^{700}$.
To avoid this, we use noise contrastive estimation (NCE)~\cite{JMLR:v13:gutmann12a}.
NCE represents the partition function as an additional learnable parameter $c$.
With noise distribution $p_n(s,a)$, NCE estimates parameters $(c, w)$ based on the following binary classification like objective function, which distinguishes whether $(s,a)$ is a sample from an empirical distribution or noise distribution,
\begin{align}
    \label{eq: reward_estimator_learning_nce}
    \begin{split}
    J&(w, c) = \\
    & \frac{1}{|B|}\sum_{(s,a) \in B} \log \frac{\exp(f_w(s,a))}{\exp(f_w(s,a)) + c \nu p_n((s,a))} \\
    & + \frac{\nu}{|B_n|}\sum_{(s,a) \in B_n}\log\frac{c \nu p_n(s,a)}{\exp(f_w(s,a)) + c \nu p_n(s,a)},
    \end{split}
\end{align}
where $B$ is a sample set from $p_D(s,a)$, $B_n$ is a sample set from $p_n(s,a)$, $|\cdot|$ is the sample size, and $\nu$ is the noise sampling ratio.

\citet{mnih:fast} empirically showed that setting $c=1$ does not affect on NCE performance if $f_w$ is a highly expressive model such as neural networks.
Therefore, it is sufficient to update only the parameter $w$.

Computing $f_{\text{min}}$ is difficult because of the large combinations of $(s,a)$.
On the other hand, NCE learns $f_w$ to assign smaller values to samples from the noise distribution.
For these reasons, instead of $f_{\text{min}}$, we use the minimum value of $f_w$ for $K$ samples from the noise distribution, i.e. $\hat{f}_{\text{min}} \coloneqq \min_{i=1,\ldots,K} \left\{f_w(s,a) \mid (s_i,a_i) \sim p_n(s,a)\right\} $.

\begin{algorithm}[tb]
\caption{proposed reward estimation method}
\label{alg: our_method}
\begin{algorithmic}[1]
    \Require expert data $D$, sampling ratio $\nu$, noise distribution $p_n$, user simulator $\mu$ 
    \ForAll{reward estimator training}
        \State Update reward estimator $p_w$ with NCE (Eq.~\eqref{eq: reward_estimator_learning_nce})
    \EndFor
    \ForAll{dialog policy training}
        \State Update dialog policy $\pi_\theta$ through interaction with $\mu$ (Eq.~\eqref{eq: obj_dialog_policy_ours})
    \EndFor
\end{algorithmic}
\end{algorithm}

In our method, the dialog policy $\pi_\theta$ is updated by maximizing the expected cumulative reward for generated dialog sessions,
\begin{equation}
    \label{eq: obj_dialog_policy_ours}
    \max_{\theta} \ \mathbb{E}_{\tau \sim \pi_\theta} \left[ \sum_{t=t_0}^{|\tau|} \gamma^{t-t_0} \hat{r}(s_t, a_t) \right].
\end{equation}
We provide the overview of our method in Algorithm~\ref{alg: our_method}.
The learning of the reward estimator is independent of the dialog policy, thus separating the learning of the dialog policy from the reward estimation.

\section{Experiment}
In this section, we present the experimental results that validate the analyses discussed in Sec.~\ref{sec: analysis_for_obj_func} and evaluate the performance of the proposed method. 
Specifically, we verify the following points:
\begin{itemize}
    \setlength{\parskip}{2mm} 
    \setlength{\itemsep}{0mm} 
    \item The turn of loop dialogs become longer.
    \item In GDPL, AL reduces the occurrence of loop dialogs.
    \item In GDPL, AL causes mode collapse.
    \item Without using AL, the same performance as GDPL can be achieved by simply suppressing loop dialogs. 
\end{itemize}

To this end, we compare the proposed method with:
\begin{itemize}
    \setlength{\parskip}{2mm} 
    \setlength{\itemsep}{0mm} 
    
    \item \textbf{GDPL}
    \item \textbf{GDPL w/o AL}: This method removes AL from GDPL, where the reward estimator is learned using the objective function~\eqref{eq: obj_reward_wo_AL}.
    First, only the reward estimator is trained.
    Next, fixing the learned reward estimator, only the dialog policy is learned.
\end{itemize}
For every method, we use Proximal Policy Optimization (PPO)~\cite{DBLP:journals/corr/SchulmanWDRK17} for learning the dialog policy.
To stabilize learning, each model is pre-trained by imitation learning on state-action pairs.
We set the max dialog turn to $T=40$.

\subsection{Implementation Details}
In our method, the dialog policy $\pi_\theta$ is a three-layer MLP with ReLU activation function.
The reward estimator $f_w$ is a two-layer MLP with softplus activation function for the output layer.
We use RMSProp~\cite{rmspror} optimizer to update the parameters of the above two models.
Hyperparameters are shown in Table~\ref{tab: hyperparameters}.
We use the author's code~\footnote{https://github.com/truthless11/GDPL} for the implementation of GDPL.
The DST, state, action in all methods, and the hyperparameters of GDPL and GDPL w/o AL are the same as in the literature~\cite{takanobu-etal-2019-guided}.

\begin{table}[tbh]
    \centering
    \begin{tabular}{c|l}
        \hline
        hyperparameter & value \\
        \hline
        decay factor $\gamma$ & $0.99$\\
        learning rate for $p_w$ and $\pi_\theta$ & $10^{-4}$\\
        noise  distribution $p_n(s,a)$ & uniform dist.\\
        sampling ratio $\nu$ & 10 \\
        approximating samples $K$ & $10^3$ \\
        loop state action reward $L$ & $3.0 \times 10^4$\\
        \hline
    \end{tabular}
    \caption{Hyperparameters for our method.}
    \label{tab: hyperparameters}
\end{table}

\subsection{Data and User Simulator}
In this work, we use MultiWOZ 1.0~\cite{budzianowski-etal-2018-multiwoz}, a multi-domain task-oriented dialog corpus consisting of 10,483 dialogs (8483/1000/1000 are train/validation/test data respectively), 7 domains, 166 system dialog acts, 66 user dialog acts.
This corpus has external databases that hold entity information.

We use the agenda-based user simulator~\cite{schatzmann-etal-2007-agenda} which receives a user goal at the start of the dialog session and probabilistically produces a user action based on predefined rules at each turn.
The user goal consists of two types of information for each domain: (1) constraints on the simulator's desired entity (e.g. price range is moderate) and (2) the requestable information that the simulator should hear from the dialog policy (e.g. restaurant address).

\subsection{Evaluation Metrics, Evaluation and Learning Methods}
We use \textit{success} and \textit{dialog turn} as evaluation metrics.
For one dialog session, \textit{success} is defined as 1 if both the booked entities entirely satisfy the constraints in the user goal and all requestable information in the user goal is given by the dialog policy and 0 otherwise.

Each method is trained for 50 epochs with 5 random seeds.
Each epoch consists of dialog sessions with a total dialog length of approximately 1024.
The same seeds are used for each method.

\begin{figure*}[h]
    \begin{tabular}{cc}
       \hspace{-3.1mm}
      \begin{minipage}[t]{0.485\hsize}
        \centering
        \includegraphics[keepaspectratio, width=\textwidth]{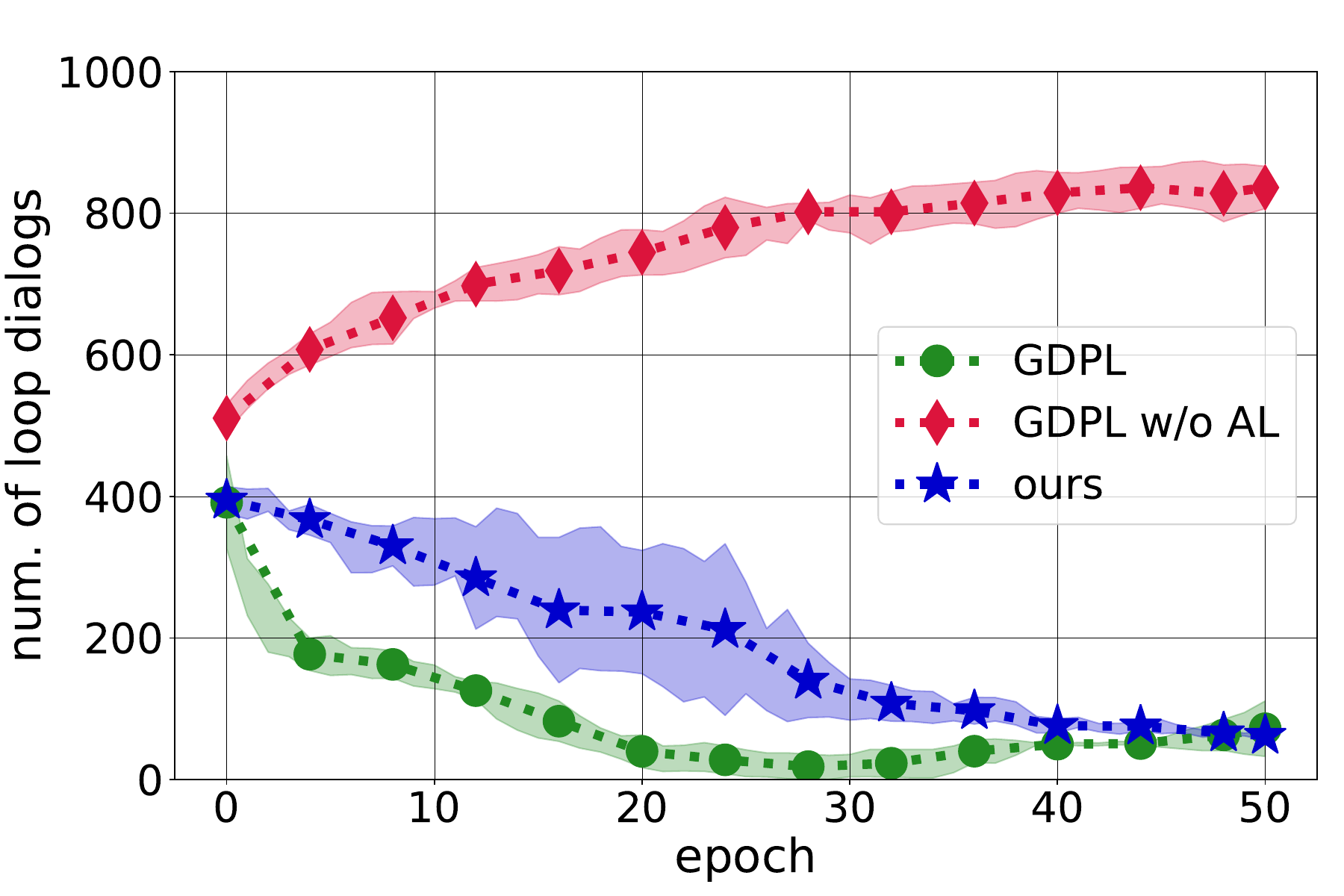}
        \caption{The number of loop dialogs in 1000 dialogs for each method.
        GDPL w/o AL removes the AL term from GDPL.
        Our method explicitly gives a large negative reward for the loop state actions without using AL.
        The dotted lines are the mean of the metric and the shades represent $\pm$ standard deviation for 5 random seeds.}
        \label{fig:num_loop_dialog}
      \end{minipage} \hspace{\columnsep}&
      \begin{minipage}[t]{0.485\hsize}
        \centering
        \includegraphics[keepaspectratio, width=\textwidth]{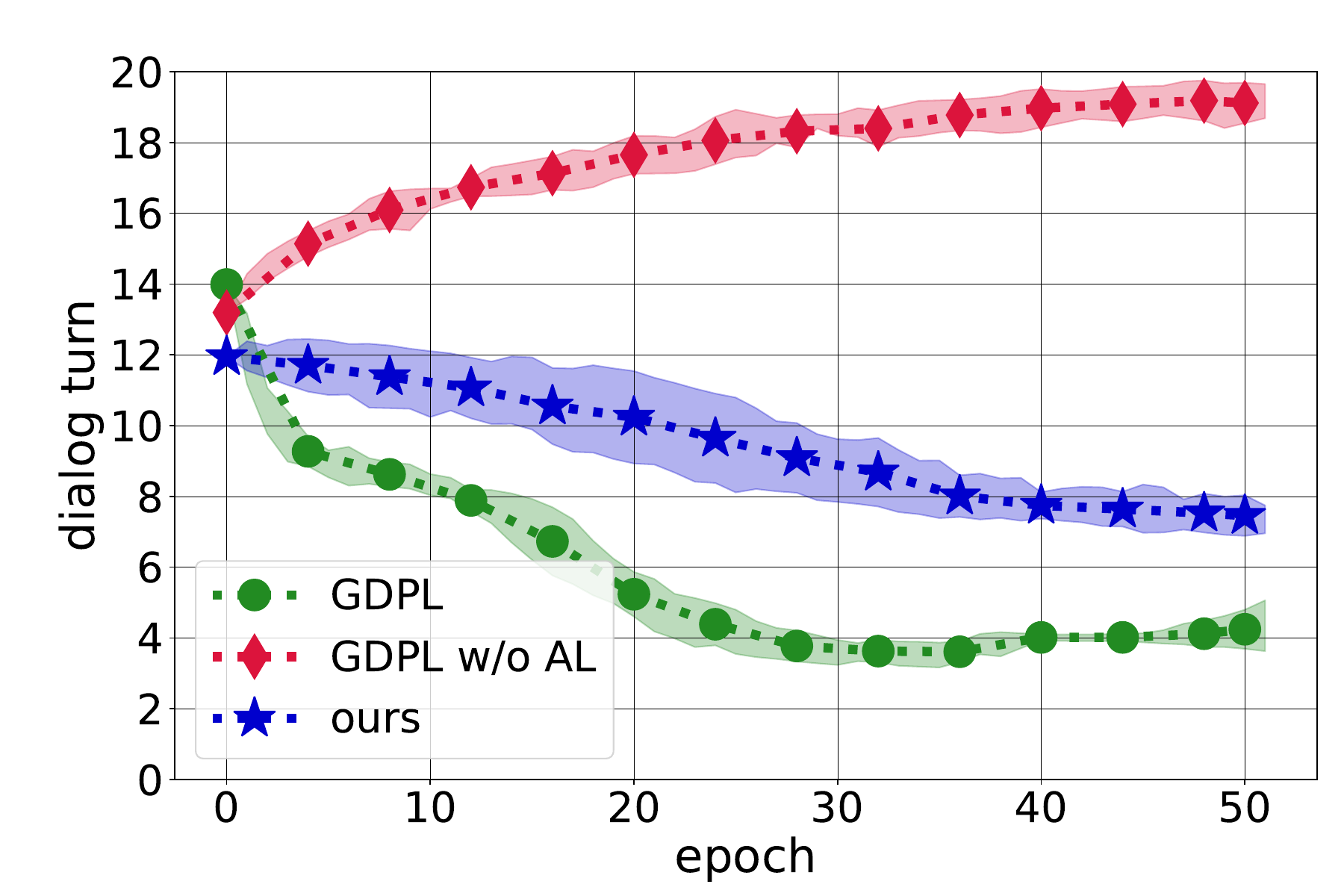}
        \caption{The transition of dialog turn for each method. 
        Dialog turn increases in GDPL w/o AL that cannot suppress loop dialogs, whereas it decreases for the other two methods that can suppress loop dialogs.}
        \label{fig:trans_of_turns}
      \end{minipage}\\

    \hspace{-3.1mm}
      \begin{minipage}[t]{0.485\hsize}
        \centering
        \includegraphics[keepaspectratio, width=\textwidth]{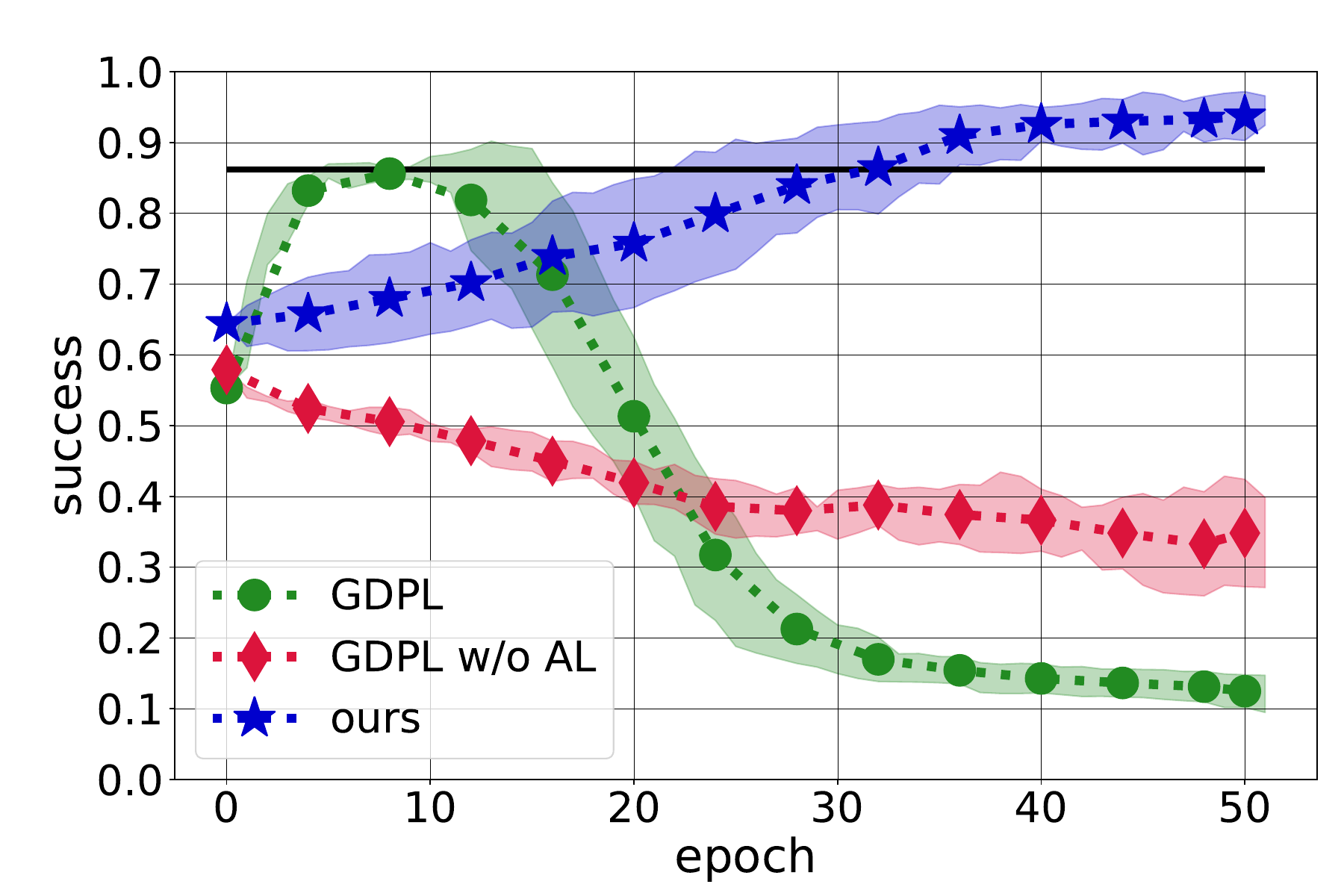}
        \caption{The transition of the rate of success in 1000 dialogs. The black solid line is the maximum value of the mean of GDPL's performance.}
        \label{fig:evaluation_metrics}
      \end{minipage} \hspace{\columnsep}&
      \begin{minipage}[t]{0.485\hsize}
        \centering
        \includegraphics[keepaspectratio, width=\textwidth]{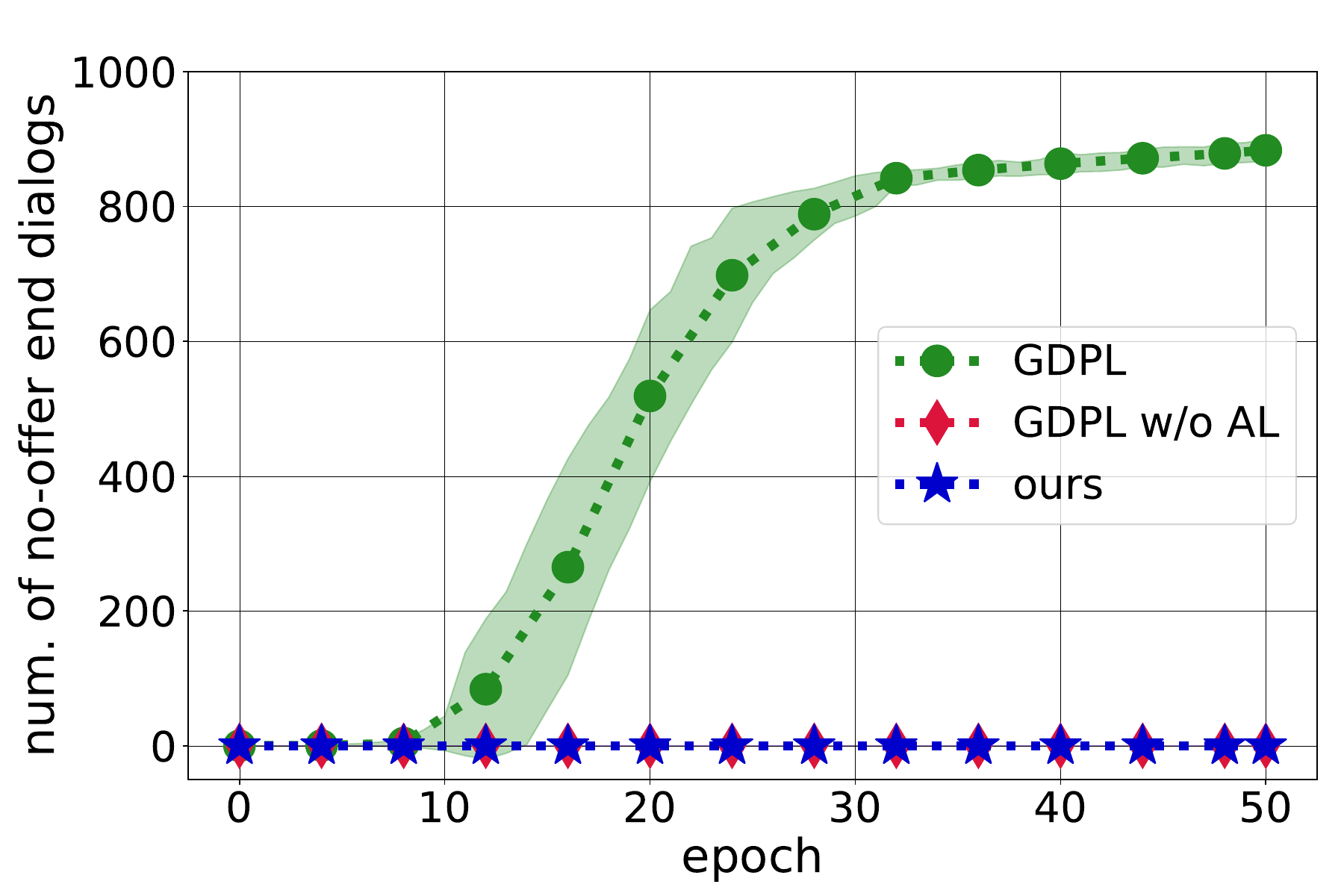}
        \caption{The transition of the number of the dialogs terminated with system actions including a dialog act \textit{no-offer}. The no-offer end dialogs have not occurred in GDPL w/o AL and our method.}
        \label{fig:trans_of_nooffer_end}
      \end{minipage}\\
      
    \end{tabular}
  \end{figure*}

At the end of each epoch, 1000 evaluation dialog sessions are generated through interaction with the user simulator, and the average of the evaluation metrics for these dialog sessions are reported.
The seeds used to generate the evaluation dialog sessions are the same for all epochs and methods.

\subsection{Results and Analyses}
Figure~\ref{fig:num_loop_dialog} shows the transitions of the number of loop dialogs for each method.
The number of loop dialogs increases in GDPL w/o AL, whereas the number of loop dialogs decreases in GDPL.
These results show that AL suppresses the occurrence of loop dialogs in GDPL.
Although our method does not use AL, it suppresses loop dialogs by providing a large negative reward for the loop state actions.

Figure~\ref{fig:trans_of_turns} suggests that the dialog turn of loop dialogs is longer.
The dialog turn decreases in GDPL and our method, where the number of loop dialogs decreases as the learning progresses, while it increases in GDPL w/o AL, where the number of loop dialogs increases.

Figure~\ref{fig:evaluation_metrics} shows the transitions of the rate of success.
GDPL and our method perform better than GDPL w/o AL.
This result suggests that the suppression of loop dialogs contributes to improved performance of dialog policies in multi-domain task-oriented dialogs.

Additionally, our method outperforms GDPL.
The performance of GDPL has worsened after epoch 10.
This is due to an increase of dialog sessions terminated with the \textit{no-offer}, which is one of the intents of the dialog acts (cf. Figure~\ref{fig:trans_of_nooffer_end}).
When the user simulator receives an action containing no-offer, it immediately terminates the dialog session.
If a dialog ends with no-offer, the success is reduced because the information that would have been heard subsequently is no longer heard, or the entities that would have been able to book subsequently are no longer able to book.

We consider that the increase in the no-offer end dialogs is mode collapse, an inherent problem in AL.
When GDPL was trained by removing dialog sessions containing no-offer from the expert data, the poor performance was avoided during the training process.
Because our method does not use AL for either learning the reward estimator or the dialog policy, we believe that it avoids the mode collapse.

\section{Conclusion and Future Work}
We identified one of the effectiveness of AL in GDPL through the analyses of its objective functions: AL suppresses loop dialogs that repeat the same state actions during one dialog session.
Furthermore, based on these analyses, we proposed a reward estimation method with NCE for DPL to address AL-specific problems.
The proposed method avoids the use of AL by explicitly assigning a large negative reward to the repeated state-action pairs in loop dialogs.
Experiments with a user simulator demonstrated that the proposed method avoids mode collapse and properly guides the learning of dialog policy by suppressing the occurrence of loop dialogs, leading to improved success rates and reduced dialog turns.

In this work, we only focus on the method that uses AL with a real-valued discriminator.
In future work, we will examine the extent to which our analyses apply to DPL methods using AL with  binary classifier discriminator (e.g. \citealp{li-etal-2020-guided}).

\bibliography{references}

\newpage
\appendix
\section{Loop dialog repeats loop state action}
\label{appendix: loop_dialog_repeat}

Loop dialog remains in the same state for $n$ turns.
This means that the actions $a_t, \ldots, a_{t+n}$ don't update the state $s_t$.
Hence, continuously selecting the actions $a_t, \ldots, a_{t+n}$  after turn $t+n$ often leads to the repetition of the loop $(s_t, a_t) \to (s_{t+n}, a_{t+n})$.
In other words, the loop state action $(s_{t+n}, a_{t+n})$ occurs multiple times.

\end{document}